\documentclass{article}
\usepackage{spconf}

\usepackage{cite}
\usepackage{amsmath,amssymb,amsfonts}
\usepackage{algorithmic}
\usepackage{graphicx}
\usepackage{textcomp}
\usepackage{xcolor}
\usepackage{array}
\usepackage{enumitem}
\usepackage{mathtools}
\usepackage{tcolorbox}
\usepackage{color}
\usepackage{theorem}
\usepackage{amssymb}
\usepackage{caption}
\usepackage{subcaption}
\usepackage{cite,hyperref}
\usepackage{cases}
\usepackage{url}
\usepackage{algorithmic}
\usepackage{algorithm}
\usepackage{tikz}
\newtheorem{definition}{Definition}
\usetikzlibrary{shapes,arrows}
\input{my_styles.sty}
\usepackage{multirow}
\usepackage{hhline}
\usepackage{algorithm,graphicx,multirow,hyperref}

\newtcolorbox{myblockt}[1]{colback=urblue!5!white,
	colframe=urblue,fonttitle=\bfseries,
	title=#1}
\newtcolorbox{myblock}{colback=urblue!5!white,
	colframe=urblue,fonttitle=\bfseries}
\def\BibTeX{{\rm B\kern-.05em{\sc i\kern-.025em b}\kern-.08em
    T\kern-.1667em\lower.7ex\hbox{E}\kern-.125emX}}

\begin{document}
\ninept

\title{Data Augmentation via Subgroup Mixup for Improving Fairness}

\name{Madeline Navarro, Camille Little, Genevera I. Allen, and Santiago Segarra 
\thanks{This work was partially supported by the NSF under award CCF-2008555. 
Research was sponsored by the Army Research Office and was accomplished under Grant Number W911NF-17-S-0002. The views and conclusions contained in this document are those of the authors and should not be interpreted as representing the official policies, either expressed or implied, of the Army Research Office or the U.S. Army or the U.S. Government. The U.S. Government is authorized to reproduce and distribute reprints for Government purposes notwithstanding any copyright notation herein.
COL acknowledges support from the NSF Graduate Research Fellowship Program under grant number 1842494.  GIA and COL acknowledge support from the JP Morgan Faculty Research Awards and NSF
DMS-2210837.
Emails:  \href{mailto:nav@rice.edu}{nav@rice.edu}, \href{mailto:col1@rice.edu}{col1@rice.edu}, \href{mailto:gallen@rice.edu}{gallen@rice.edu},\href{mailto:segarra@rice.edu}{segarra@rice.edu}}}
\address{Electrical and Computer Engineering, Rice University, USA}
\maketitle

\begin{abstract}
In this work, we propose \emph{data augmentation via pairwise mixup across subgroups to improve group fairness}.
Many real-world applications of machine learning systems exhibit biases across certain groups due to under-representation or training data that reflects societal biases.  
Inspired by the successes of mixup for improving classification performance, we develop a pairwise mixup scheme to augment training data and encourage fair and accurate decision boundaries for all subgroups.
Data augmentation for group fairness allows us to add new samples of underrepresented groups to balance subpopulations.
Furthermore, our method allows us to use the generalization ability of mixup to improve both fairness and accuracy.
We compare our proposed mixup to existing data augmentation and bias mitigation approaches on both synthetic simulations and real-world benchmark fair classification data, demonstrating that we are able to achieve fair outcomes with robust if not improved accuracy.
\end{abstract}

\begin{keywords}
    Fairness, data augmentation, mixup, group fairness
\end{keywords}

\section{Introduction}
\label{s:intro}

Machine learning (ML) algorithms are now widely employed to guide crucial decision-making processes in various high-stakes contexts, spanning sectors like finance, criminal justice, and healthcare \cite{Pessach:2022}. 
However, alongside this pervasive adoption of ML, concerns have emerged regarding the potential for bias and discriminatory decision-making, particularly concerning attributes such as race, gender, age, and other sensitive factors. 

Several methods have attempted to mitigate bias in ML systems. 
Specifically, some methods alter the observed data by removing features or rebalancing the data with respect to certain protected attributes to eliminate underlying discrimination  \cite{Kamiran:2012, Feldman:2015,Calmon:2017, Zemel:2013}. 
Other methods directly alter the ML model by adding a fairness-aware constraint or penalty to improve fair model behavior \cite{Donini:2018, Do:2022}. 
While several existing methods have been shown to improve fairness, they often have several limitations. 
For one, many techniques cannot be used beyond a specific classifier or family of classifiers. 
Additionally, techniques that completely remove certain features tend to suffer from a major decrease in model performance in terms of accuracy. 
Lastly, several of these methods use explicit regularization to minimize bias for models trained on a training set and, in many cases, do not generalize well to out-of-sample data \cite{cotter:2019}. 

We require an approach to improve model fairness that generalizes to unseen data, does not sacrifice accuracy greatly, and does not restrict model capacity.
Data augmentation is an implicit regularization method that is well known to improve model generalization for prediction tasks~\cite{poggio:2017}. 
Mixup is one such data augmentation method that trains a model on convex combinations of pairs of existing samples and their labels \cite{zhang_mixup:2018}. 
While the advantages of mixup for model prediction performance are empirically and theoretically well studied, its application to fairness is underexplored, motivating us to apply mixup for mitigating bias while maintaining accuracy on out-of-sample data. 
Few existing works have investigated the effect of mixup and other data augmentation methods on model properties other than prediction accuracy~\cite{zhang2022when}, and comparatively little has been explored for fair data augmentation~\cite{han2022umix,chuang:2021,sharma2020data}.

We propose a simple and flexible pairwise linear mixup approach for fair data augmentation.
Traditional linear mixup trains on pairs of samples from different classes to improve model accuracy, and we extend beyond this specific task to improve fairness and accuracy by applying linear mixup between pairs of samples belonging to different classes and different groups. 
The proposed scheme can transform model behavior by interpolating between samples and their nearest neighbors in carefully chosen different subpopulations, the choice of which is adaptable for improving fairness under different data distributions and group fairness definitions.

Our contributions are as follows.
\begin{itemize}[labelwidth=1em,leftmargin =\dimexpr\labelwidth+\labelsep\relax]
\item[(1)] We perform data augmentation via the efficient and effective pairwise mixup to improve the fairness and accuracy of model predictions;
\item[(2)] We propose a pairwise mixup scheme across dataset subpopulations that is adaptable to different types and levels of bias; and 
\item[(3)] We empirically prove when mixup mitigates different settings of bias with synthetic simulations, and we demonstrate the viability of our approach on real-world benchmark fairness datasets.
\end{itemize}


The remainder of this work is presented as follows.
Section~\ref{s:background} provides the necessary background on group fairness and mixup for data augmentation.
We present our proposed fair mixup method in Section~\ref{s:method}.
In Section \ref{s:results}, we demonstrate multiple scenarios in which mixup aids fairness, including an essential benchmark dataset, where we use mixup schemes from our proposed approach.
Finally, we conclude in Section \ref{s:conclusion} with a discussion on the consequences of this work and its future directions.

\begin{algorithm}
    \caption{Fair SubGroup (SG) Mixup}

    \textbf{Input} ~ $\ccalD=\{(X^{(i)},Y^{(i)},Z^{(i)})\}_{i=1}^T$, ~~ $\ccalD'=\varnothing$ ~~ $K,T'\in\mathbb{N}$,

    \quad\qquad $\alpha>0$, ~~ $(Y_s, Z_s), (Y_t, Z_t)\in \ccalY\times\ccalZ$ \\
    \textbf{while $|\ccalD'|<T'$:}
    \begin{itemize}
        \item Sample $i\in\{1,2,\dots,T\}$ such that $(Y^{(i)},Z^{(i)})=(Y_s,Z_s)$
        \item Obtain $K$ nearest neighbors $\{X^{({j_k})}\}_{k=1}^K$ of $X_i$ such that $(Y^{({j_k})},Z^{({j_k})})=(Y_t,Z_t)$ for every $k\in\{1,2,\dots,K\}$
        \item Sample mixup parameter $\lambda\sim \mathrm{Beta}(\alpha,\alpha)$
        \item \textbf{for every $k\in\{1,2,\dots,K\}$:}
        \begin{subequations}\label{eq:mix}
            \begin{alignat}{3}&
                \!\!\!\!\!\!\!\!\!\!\!\!\!\!\!\!
                X_\mathrm{new}^{(k)} = (1-\lambda)X^{(i)} + \lambda X^{({j_k})}
                &\label{eq:xmix}\\&
                \!\!\!\!\!\!\!\!\!\!\!\!\!\!\!\!
                Y_\mathrm{new}^{(k)} = \mathbb{I}\{(1-\lambda)Y^{(i)} + \lambda Y^{({j_k})}\geq 1/2\}
                &\label{eq:ymix}\\&
                \!\!\!\!\!\!\!\!\!\!\!\!\!\!\!\!
                Z_\mathrm{new}^{(k)} = \mathbb{I}\{(1-\lambda)Z^{(i)} + \lambda Z^{({j_k})}\geq 1/2\}
                \label{eq:zmix}
            &\end{alignat}
        \end{subequations}
        \item Update $\ccalD'$ as $\ccalD'\cup\{(X_\mathrm{new}^{(k)},Y_\mathrm{new}^{(k)},Z_\mathrm{new}^{(k)}\}_{k=1}^K$
    \end{itemize}
    \textbf{Output} ~ Augmented dataset $\ccalD\cup \ccalD'$
\label{alg:navmf}
\end{algorithm}

\section{Background}
\label{s:background}

\noindent\textbf{Group fairness.}
We present the notion of group fairness, where a prediction model is considered fair across groups if model treatment is equivalent with respect to each group separately. 
Group fairness requires that predictions be equal across all groups \cite{Feldman:2015 ,Hardt:2016}, which is formalized in the fairness definition provided in \cite{Hardt:2016}.
\begin{definition}[Demographic Parity~\cite{Hardt:2016}] A predictive model $f$ satisfies demographic parity (DP) if predictions $\hat{Y}=f(X)$ and the protected variable $Z$ are independent. In other words, $\mathbb{P}[\hat{Y}=y]$ is equal for all values of $Z$, that is, $\mathbb{P}[\hat{Y} = y] = \mathbb{P}[\hat{Y} = y| Z= z]$.
\end{definition}
%
In this work, we consider data belonging to binary groups.
That is, for a given set of $T$ data samples, the $i$-th sample is assigned a group identity $Z_i\in\{0,1\}$.
Under this setting, we introduce DP gap as 
\begin{equation}\label{eq:dp}
\Delta \mathrm{DP} := 
 \frac{1}{|G_0|}\sum_{i \in G_0} \hat{Y}^{(i)} - \frac{1}{|G_1|}\sum_{i \in G_1} \hat{Y}^{(i)},
\end{equation}
where $\hat{Y}^{(i)}$ denotes the $i$-th prediction and $G_z=\{j:Z_j=z\}$ for $z\in\{0,1\}$. 
Note that \eqref{eq:dp} is amenable to regression tasks, to which our method can be adapted under slight modifications~\cite{yao2022cmixup}.
When $\Delta\mathrm{DP}=0$, we have achieved DP; otherwise, there is a \emph{gap in DP}, which bias mitigation methods aim to reduce.
While we highlight and implement DP in this paper, our method can be easily extended to other group fairness definitions, such as equalized odds~\cite{Hardt:2016}. 

\medskip

\noindent\textbf{Mixup data augmentation.}
Mixup for data augmentation is a cheap but effective method that generates data from convex combinations of existing pairs of labeled samples~\cite{zhang_mixup:2018,verma2019manifold,yun2019cutmix}. 
However, the primary use of mixup is to improve model prediction accuracy with far less exploration on its ability to improve other metrics~\cite{zhang2022when,han2022umix,chuang:2021}.
While authors in~\cite{han2022umix} apply mixup in the case where group membership frequency differs between training and testing distributions, they emphasize mitigating error and do not address model treatment of the groups.
The closest work to our own is \cite{chuang:2021}, which also demonstrates the value of pairwise linear mixup for improving group fairness.
However, their approach is fundamentally different from ours as they apply mixup samples solely for regularization, whereas we explicitly augment the dataset with mixed samples chosen to improve fairness. 
Both their approach and ours achieve favorable bias mitigation, although our data augmentation approach needs no restriction on the model nor an explicit regularization term.

\section{Methodology}
\label{s:method}

Consider the labeled dataset $\ccalD=\{(X^{(i)},Y^{(i)},Z^{(i)})\}_{i=1}^T$ containing $T$ samples, where a sample $(X,Y,Z)\in\ccalD$ consists of a set of $d\in\mathbb{N}$ features $X\in\ccalX\subset\mathbb{R}^d$, a label $Y\in\ccalY:=\{0,1\}$, and a protected attribute $Z\in\ccalZ:=\{0,1\}$.
The presence of $Z$ partitions the dataset into two groups, and we may further divide the dataset into \emph{subgroups} according to both class and group assignment $(Y,Z)$.
Our goal is to generate synthetic data $\ccalD'=\{(X^{(i)}_\mathrm{new},Y^{(i)}_\mathrm{new},Z^{(i)}_\mathrm{new})\}_{i=1}^{T'}$ via pairwise linear mixup~\cite{zhang_mixup:2018} such that when we fit a classifier model $f:\ccalX\rightarrow\ccalY$ to the augmented dataset $\ccalD\cup\ccalD'$, the resultant model is fair across groups in $\ccalZ$.
In particular, we aim to encourage DP, where the distribution of predicted labels is approximately equivalent for each group.
To this end, we use mixup to generate new samples that reduce DP gap $\Delta\mathrm{DP}$ when added to the existing training data.
Our proposed mixup process is presented in Algorithm \ref{alg:navmf}, and we elaborate on our approach in the sequel.

The crux of Algorithm~\ref{alg:navmf} is interpolation between \emph{subgroups}, from a source subgroup, $(X_s, Y_s, Z_s)$, to a target subgroup, \\$(X_t, Y_t, Z_t)$.
Unlike traditional linear mixup, which interpolates between classes for model accuracy, we wish to maintain or improve accuracy while encouraging consistent treatment across groups. 
We thus perform mixup between samples belonging to different subgroups, that is, we apply \eqref{eq:mix} for a given pair $(X^{(i)},Y^{(i)},Z^{(i)})$, $(X^{(j)},Y^{(j)},Z^{(j)})$ with $(Y^{(i)},Z^{(i)})\neq(Y^{(j)},Z^{(j)})$.
The mixup parameter $\lambda\in[0,1]$ dictates to which of the original samples the new data is closer, and $\mathbb{I}\{\cdot\}$ denotes the indicator function. 
The resultant samples populate the convex hull across the subgroups, encouraging model invariance between them.

\begin{table}[!t]
    \footnotesize
    \centering
    \begin{tabular}{c | c c}
         & $Y=0$ & $Y=1$ \\ \hline
         $Z=0$ & $T_{00}/(T_{00}+T_{10})$ & $T_{10}/(T_{00}+T_{10})$ \\
         $Z=1$ & $T_{01}/(T_{01}+T_{11})$ & $T_{11}/(T_{01}+T_{11})$ \\
    \end{tabular}
    \caption{\small{Subgroup proportions, normalized by samples per group.}}
    \label{tab:sg_props}
    \vspace{-2mm}
\end{table}

\begin{figure*}[th!]
    \centering
    \begin{minipage}[c]{.27\textwidth}
		\includegraphics[width=\textwidth]{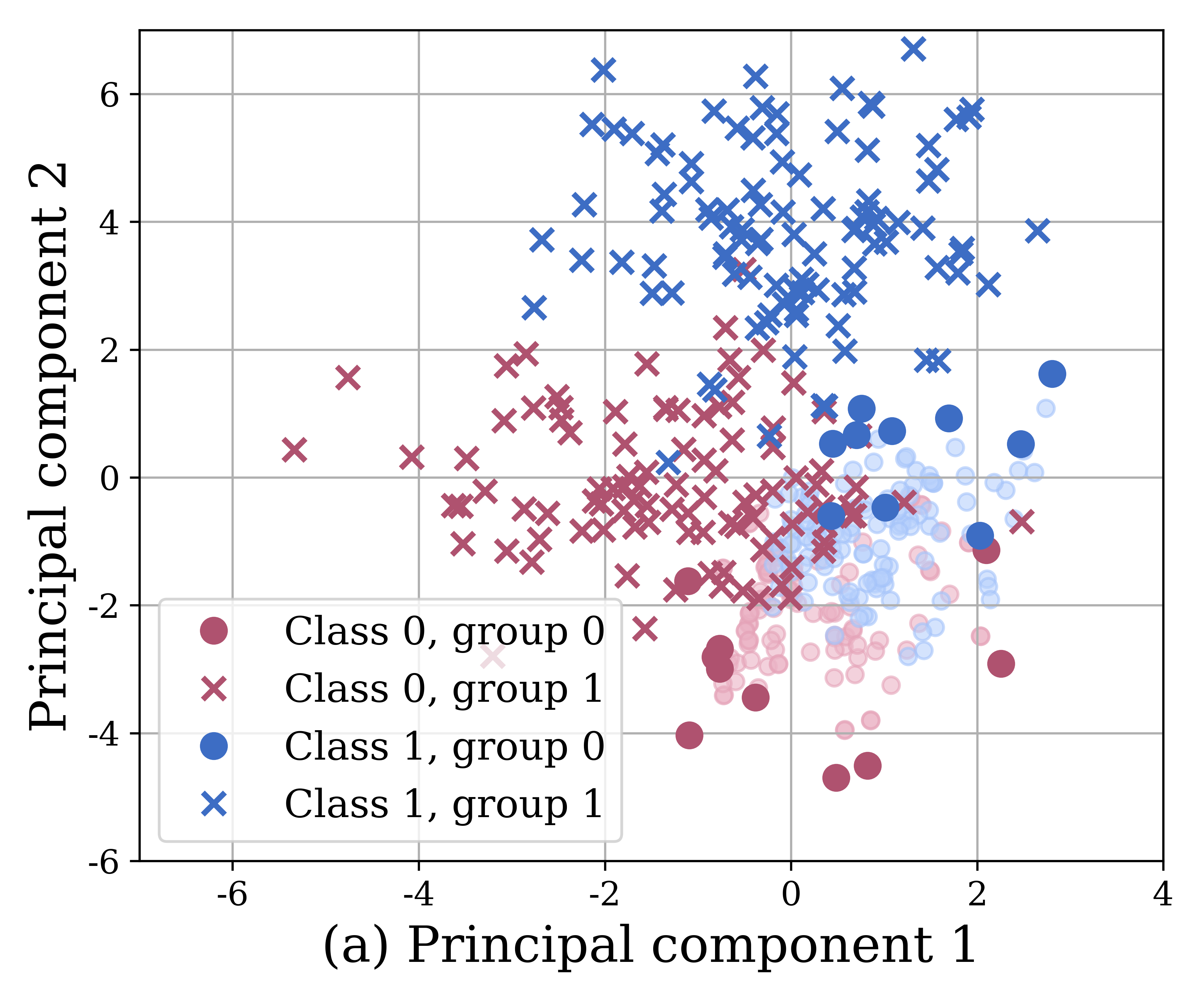}
	\end{minipage}
	\begin{minipage}[c]{.27\textwidth}
		\includegraphics[width=\textwidth]{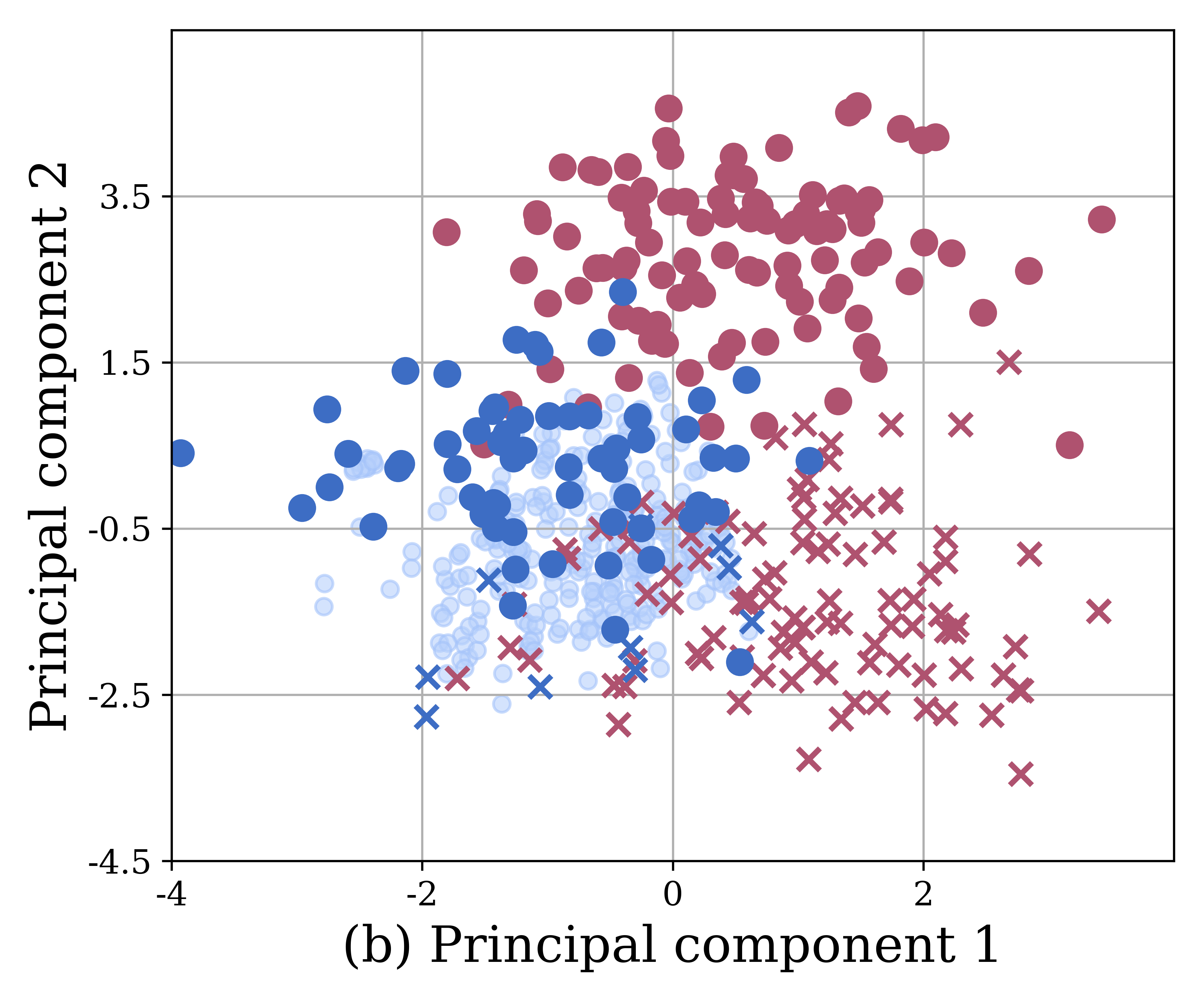}
	\end{minipage}
	\begin{minipage}[c]{.27\textwidth}
		\includegraphics[width=\textwidth]{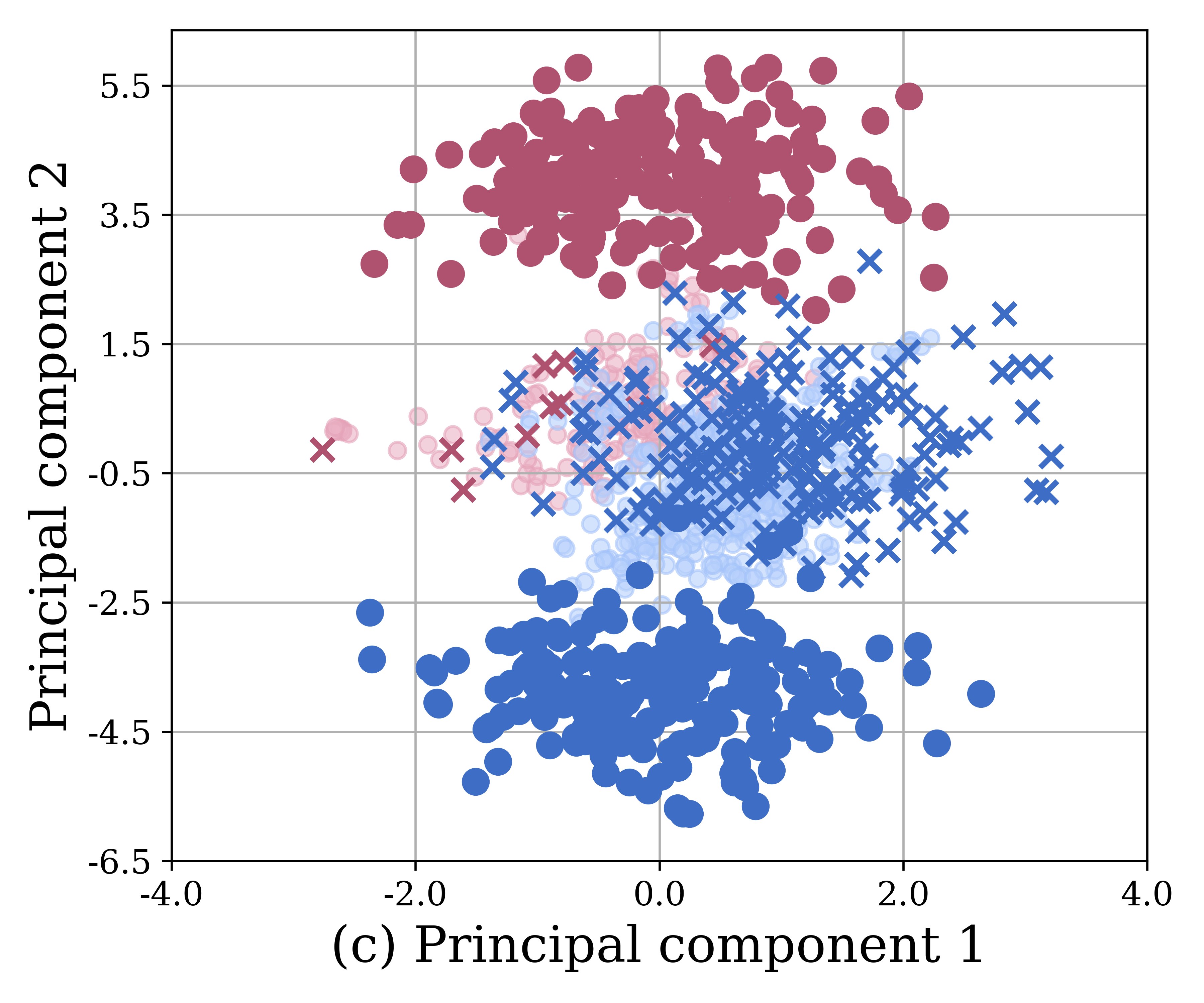}
	\end{minipage}
 
    \begin{minipage}[c]{.27\textwidth}
		\includegraphics[width=\textwidth]{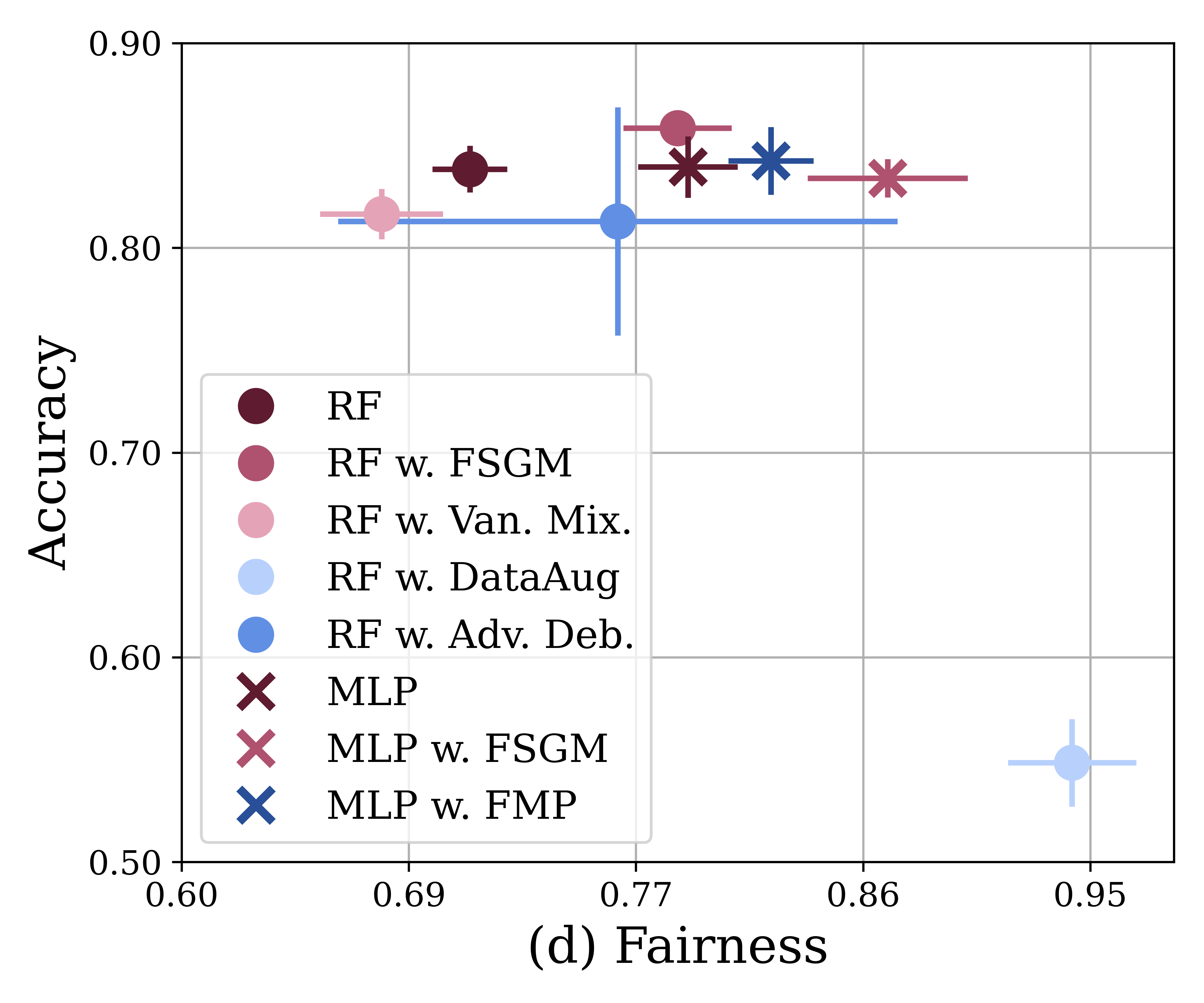}
	\end{minipage}
	\begin{minipage}[c]{.27\textwidth}
		\includegraphics[width=\textwidth]{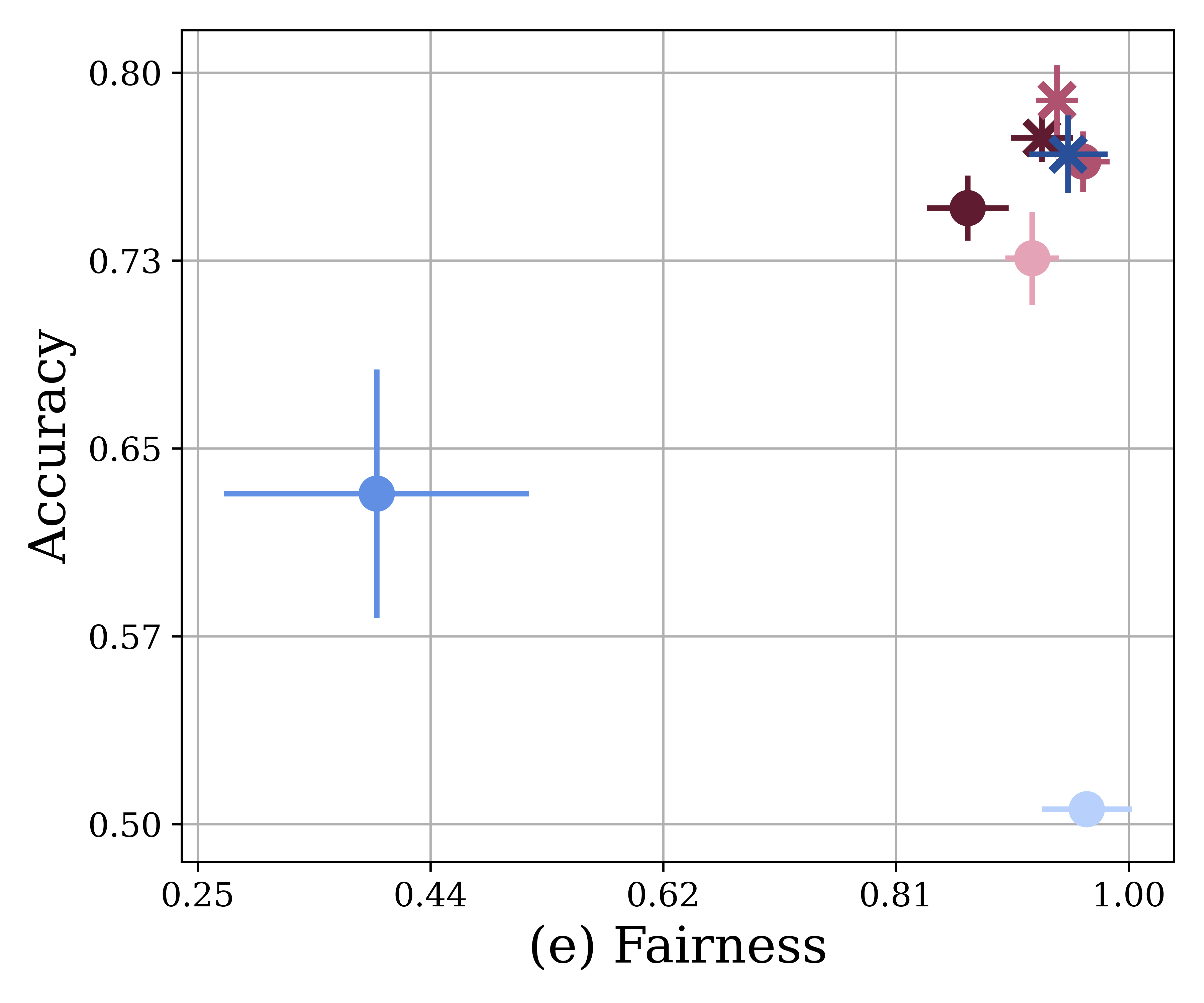}
	\end{minipage}
	\begin{minipage}[c]{.27\textwidth}
		\includegraphics[width=\textwidth]{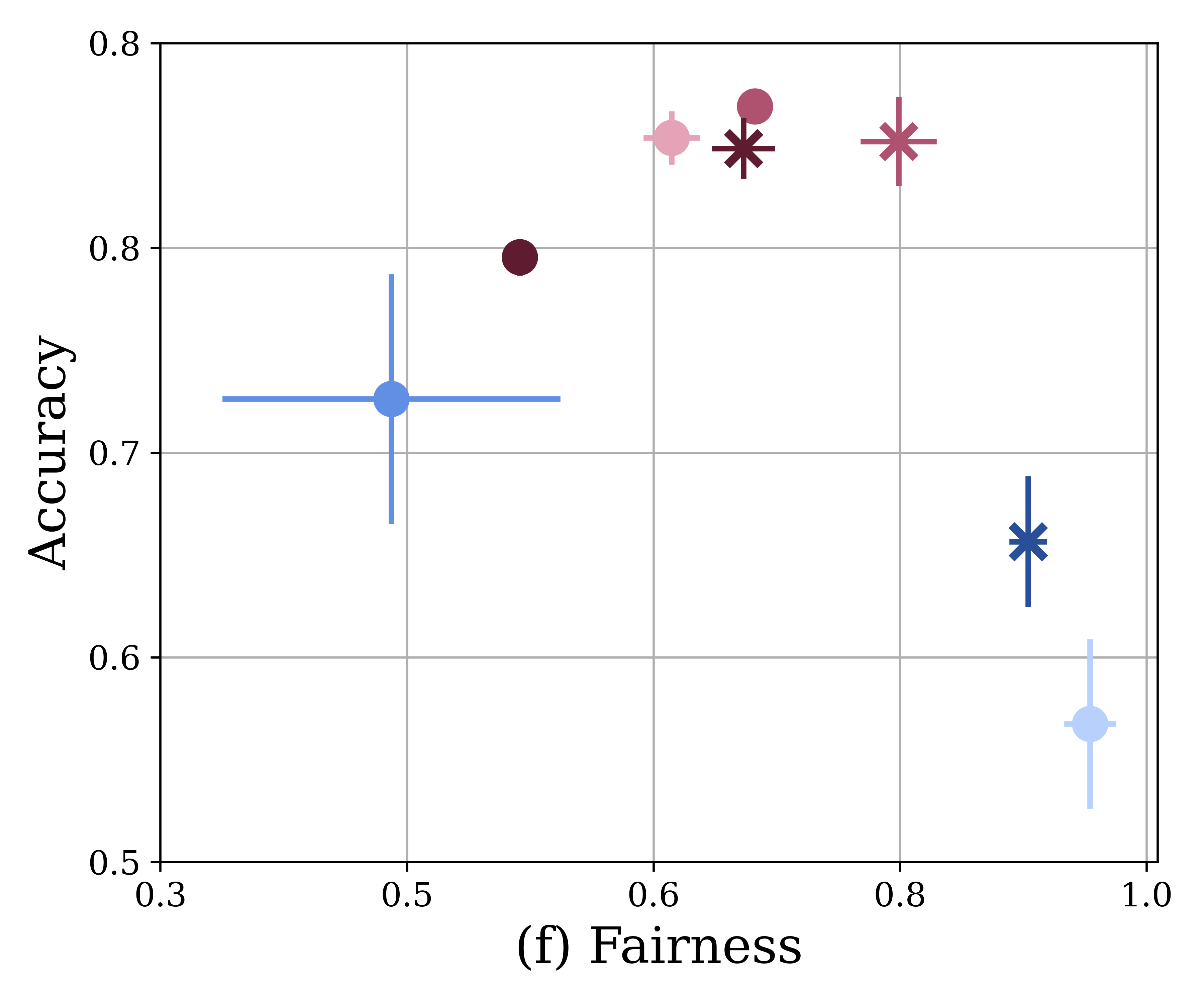}
	\end{minipage}
    \caption{\small{Comparison of Fair SG Mixup for unbalanced subgroups. (a) Labeled data from~\eqref{eq:nbdata} with unbalanced groups. (b) Labeled data from~\eqref{eq:nbdata} with unbalanced classes. (c) Labeled data from~\eqref{eq:nbdata} with heavily underrepresented subgroup. (d) Comparison of bias mitigation and data augmentation methods for unbalanced groups. Fair SG mixup interpolates between the same underrepresented group and across classes. (e) Comparison of bias mitigation and data augmentation methods for unbalanced classes. Fair SG mixup interpolates between the same underrepresented class and across groups. (f) Comparison of bias mitigation and data augmentation methods for one underrepresented subgroup. Fair SG mixup interpolates both between classes and between groups.}}
\label{f:mfres}
\end{figure*}

\subsection{Subgroups for fair mixup}
\label{ss:sg_bias}

We discuss the versatility of Algorithm~\ref{alg:navmf} for improving model fairness in different settings.
Our approach mitigates bias caused by both unbalanced subgroups and subgroup distribution shifts via the choice of $(Y_s,Z_s)$ and $(Y_t,Z_t)$.
To see this, first let $T_{yz}$ denote the number of samples in $\ccalD$ belonging to the subgroup $(y,z)\in\ccalY\times\ccalZ$.
Proportions of samples per subgroup are shown in Table~\ref{tab:sg_props}.
To achieve DP, rows in Table~\ref{tab:sg_props} must be equivalent.

\medskip 

\noindent\textbf{Unbalanced subgroups.}
First, we consider the choice of the source subgroup $(Y_s,Z_s)$ in Algorithm~\ref{alg:navmf} to mitigate bias from unbalanced subgroups.
An obvious advantage of data augmentation is adding new samples to a row of Table~\ref{tab:sg_props} to enforce DP.
However, we reap further benefits of mixup beyond only balancing subgroups.

Algorithm~\ref{alg:navmf} returns samples from $(Y_s,Z_s)$ that are perturbed toward $(Y_t,Z_t)$.
More specifically, we interpolate each sample from $(Y_s,Z_s)$ with its nearest neighbors with respect to the features $X$ in the target subgroup $(Y_t,Z_t)$.
The resultant vicinal distribution is known to improve generalization by encouraging linear transitions between subpopulations~\cite{zhang_mixup:2018}.
Consider choosing the source $(Y_s,Z_s)$ to be an underrepresented subgroup, that is, with the smallest entry in Table~\ref{tab:sg_props}.
In this setting, we promote smooth behavior near the underrepresented subgroup in the direction of the target $(Y_t,Z_t)$.
This encourages the model to treat samples from the underrepresented subgroup more similarly to those from the target~\cite{zhang_mixup:2018}.
Thus, the proportions in Table~\ref{tab:sg_props} provide a natural choice of $(Y_s,Z_s)$ that mitigates bias from imbalanced data.

\medskip 

\noindent\textbf{Subgroup distribution shifts.}
Appropriate choice of the target subgroup $(Y_t,Z_t)$ in Algorithm~\ref{alg:navmf} ameliorates bias from distribution shifts.
Such bias may be invisible in Table~\ref{tab:sg_props}, but our mixup scheme permits not only balancing subgroup proportions but also adjusting model behavior by our choice of $(Y_t,Z_t)$.
Model predictions may differ greatly across groups in the presence of a large mean shift dependent on $Z$.
In this case, consider choosing $(Y_t,Z_t)=(Y_s,1-Z_s)$, resulting in mixup within the same class but across groups.
We generate samples belonging to class $Y_s$ that promote constant model predictions in between groups, as is the goal of DP.

Alternatively, we may let $(Y_t,Z_t)=(1-Y_s,Z_s)$, interpolating between samples from the same group but across classes.
Due to the indicator function in \eqref{eq:ymix}, mixup samples encourage a stricter decision boundary in between classes.
More specifically, for any new sample $(X_\mathrm{new},Y_\mathrm{new},Z_\mathrm{new})$ and features $X_s$,$X_t$ corresponding respectively to samples from the source and target subgroups, we have that $Y_\mathrm{new}=Y_s$ when $\|X_\mathrm{new}-X_s\|_2 < \|X_\mathrm{new}-X_t\|_2$, and $Y_\mathrm{new}=1-Y_s$ otherwise.
The new samples then encourage certainty in model predictions \emph{per group}. 
This not only mitigates bias from class-dependent distribution shifts but also improves accuracy since we specify model predictions separately for each group.
We then have a practical way to choose the target subgroup, where we measure empirical distribution shifts across subgroups and select $(Y_t,Z_t)$ with respect to the source $(Y_s,Z_s)$ accordingly.



\section{Results}
\label{s:results}

We illustrate the merits of subgroup mixup for fairness in multiple scenarios.
We first show when subgroup mixup improves classifier fairness in the presence of different forms of bias as described in Section~\ref{ss:sg_bias}.
We then demonstrate the value of mixup for fair data augmentation on a benchmark real-world fair classification dataset. 

\medskip

\subsection{Synthetic simulations}
We first empirically demonstrate bias due to subgroup imbalance and distribution shift.
We generate $T$ labeled data samples $(X,Y,Z)$ from a conditional Gaussian model, where the features $X\in\mathbb{R}^d$ for $d=10$ follow a Gaussian distribution
\begin{equation}\label{eq:nbdata}
    \mathbb{P}[X | Y,Z] = \prod_{i=1}^d \frac{1}{\sqrt{2\pi}}
    \exp\left\{
    -\frac{1}{2} \left( 
    (X_i-B_i(Y)-C_i(Z)
    \right)^2 \right\},
\end{equation}
where $B:\ccalY\rightarrow\mathbb{R}^d$ denotes the mean shift due to the class $Y$ and $C:\ccalZ\rightarrow\mathbb{R}^d$ the shift due to group $Z$.

We compare our proposed method to data augmentation and bias mitigation baselines. 
In particular, our approach for Fair SG Mixup in Algorithm~\ref{alg:navmf}, denoted ``FSGM'', is compared with (i) the original dataset; (ii) ``Van. Mix.'', vanilla mixup between random sample pairs from different classes~\cite{zhang_mixup:2018}; (iii) ``DataAug'', fair data augmentation in~\cite{sharma2020data} that adds copies of existing samples while swapping the group label $Z$; (iv) ``Adv. Deb.'', which applies adversarial learning for fairness by learning how to prevent an adversary from predicting the protected attribute; and (v) ``FMP'', the fair mixup penalty applied in~\cite{chuang:2021}, which does not explicitly perform data augmentation but instead regularizes model behavior between groups.
For all data augmentation methods, we augment the dataset to $2T$ samples, and for remaining baselines, including the original dataset, we apply bootstrapping so the training data size matches that of the data augmentation methods.

For the following synthetic simulations, we train either a Random Forest classifier model, denoted ``RF'', or a multilayer perceptron (MLP), denoted ``MLP'', on five independent synthetic datasets of the form in~\eqref{eq:nbdata}.
For ``Van. Mix.'' and ``FSGM'', we sample the mixup parameter $\lambda\sim\mathrm{Beta}(\alpha,\alpha)$, the usual choice for mixup~\cite{zhang_mixup:2018}.
For each simulation, we apply the value of $\alpha>0$ that achieves the highest sum of accuracy and fairness.

We show simulation results in Fig.~\ref{f:mfres} for different settings of bias.
Fig.~\ref{f:mfres}a to c visualizes the synthetic data in each case, where the augmented data resulting from our Fair SG Mixup is shown in lighter colors with smaller markers.
We present in Fig.~\ref{f:mfres}d to f the model accuracy versus fairness of predictions on an unseen test dataset~\cite{little2022to}, where fairness corresponds to $1-\Delta\mathrm{DP}$ from~\eqref{eq:dp}.

\medskip

\begin{figure}[t!]
    \centering
		\includegraphics[width=.38\textwidth]{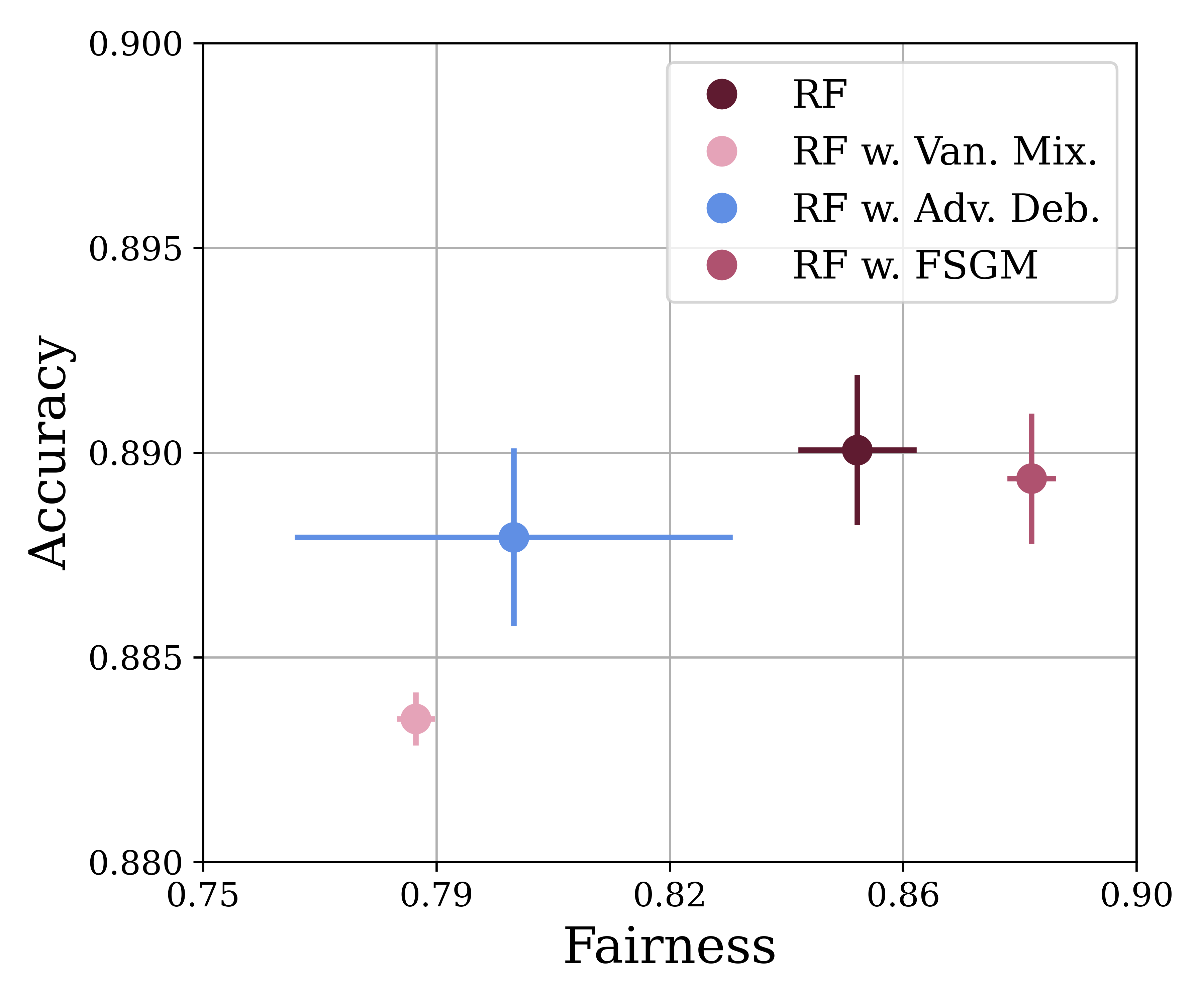}
    \caption{\small{Comparison of Fair SG Mixup with bias mitigation and data augmentation methods on the Law School Admission Bar Passage dataset with race as the protected attribute \cite{law_data}. Fair SG Mixup interpolates between samples from the same group and across classes. }}
\label{f:law}
\end{figure}

\noindent\textbf{Unbalanced groups.}
We first consider the case of unbalanced groups, visualized in Fig.~\ref{f:mfres}a, where $T_{00}=T_{10}=10$ and $T_{01}=T_{11}=100$.
In addition to bias due to the mismatch in group sizes, we shift subgroup distributions with $B(Y)=-B(1-Y)$ and $C(Z)=-C(1-Z)$.
In the presence of an underrepresented group, we choose $(Y_s,Z_s)\in\{(0,0),(1,0)\}$ with $(Y_t,Z_t)=(1-Y_s,Z_s)$.

In Fig.~\ref{f:mfres}d, applying our mixup approach ``FSGM'' with Random Forest ``RF'' not only improves model fairness but also maintains and even improves prediction accuracy.
Indeed, ``Adv. Deb.'' exhibits a comparatively unstable increase in fairness but a decrease in accuracy.
Moreover, ``Van. Mix.'' does not consider fair treatment of groups, thus for underrepresented groups, it does not outperform the original dataset in fairness.
We note that in all scenarios, ``DataAug'' achieves a high fairness result since it selectively bootstraps samples to create fairer data, but the accuracy consistently decreases to around 50\%, hence it is not viable for practical implementation in these settings.
Our method ``FSGM'' and the fair mixup penalty ``FMP'' both improve ``MLP'' model fairness with little decrease in accuracy, with a greater increase in fairness for ``FSGM''.
We thus demonstrate that mixup across classes bolsters model performance for an underrepresented group.

\medskip

\noindent\textbf{Unbalanced class.}
We next evaluate mixup in the presence of unbalanced classes, shown in Fig.~\ref{f:mfres}b and e. 
Here, we let $T_{00}=T_{01}=100$, and $T_{10}=60$ while $T_{11}=10$.
In this setting, we let $(Y_s,Z_s)=\{(1,0),(1,1)\}$ and $(Y_t,Z_t)=(Y_s,1-Z_s)$.

First, we observe that implicit regularization methods demonstrate superior performance in terms of both fairness and accuracy.
The bias mitigation method ``Adv. Deb.'' is far outperformed by all other methods, and all methods besides ``Adv. Deb.'' rival ``DataAug'' in terms of model fairness.
For ``RF'', our method ``FSGM'' again exhibits the greatest increase in both accuracy and fairness.
While ``MLP'' achieves similar fairness with and without ``FSGM'' or ``FMP'', our approach enjoys greater accuracy.

\medskip

\noindent\textbf{Underrepresented subgroup.}
Finally, we consider the case where one subgroup is heavily underrepresented, with $T_{00}=T_{01}=T_{11}=200$ while $T_{10}=10$, shown in Fig.~\ref{f:mfres}c.
The groups are balanced for $Y=0$ but unbalanced for $Y=1$, resulting in dissimilar rows of Table~\ref{tab:sg_props}.
We also introduce additional bias, where we let the angle between distribution shifts $B(Y)$ and $C(Z)$ be $\pi/6$, resulting in increased correlation between the classes and groups.
In this setting, we select the underrepresented subgroup $(1,0)$ as the source $(Y_s,Z_s)$, and for the target $(Y_t,Z_t)$ we choose $(1,1)$ and $(0,0)$. 
Even in the presence of both unbalanced subgroups and subgroup distribution shifts that contribute bias, we still find that our proposed ``FSGM'' improves results from the original dataset and is superior to ``Adv. Deb.'' and ``Van. Mix.''.
Furthermore, for the ``MLP'' model, ``FSGM'' maintains accuracy while increasing fairness, while ``FMP'' experiences a drop in accuracy despite achieving a higher fairness.

\subsection{Case Study on Real-Data }
We next examine the performance of our fair data augmentation method on the popular Law School Admission Bar Passage benchmark dataset \cite{law_data}. 
The dataset, whose target labels $Y$ state whether or not a student will pass the bar, has 20,100 students and 7 features. 
We use Race as the protected attribute $Z$ encoded as white (1) or nonwhite (0). 
In Fig.~\ref{f:law}, we compare our approach ``FSGM'' along with others using Random Forest ``RF''. 
We omit ``DataAug'' due to its inferior performance with respect to accuracy, as demonstrated in Fig.~\ref{f:mfres}. 
We choose $(Y_s,Z_s)=\{(1,0),(0,0)\}$ and $(Y_t,Z_t)=(1-Y_s,Z_s)$.
We again observe that ``FSGM'' achieves a higher fairness than all other methods, notably superior to bias mitigation method ``Adv. Deb.'', while experiencing a minor drop in accuracy, yet still outperforming traditional mixup ``Van. Mix.''.
This validates our approach as a robust data augmentation method for improving both accuracy and fairness in real-world situations using domain-agnostic convex combinations of existing data.

\section{Conclusion}
\label{s:conclusion}

In this work, we proposed a linear pairwise mixup scheme for augmenting data with samples to improve group fairness.
We extend the traditional mixup approach of interpolating between samples belonging to different classes to interpolation across subgroups, where samples may belong to different classes or groups, which correspond to values of a discrete protected attribute.
We empirically demonstrated that our approach is adaptable and viable in many scenarios of bias present in training data.
Crucially, we showed that we can not only improve fairness beyond existing bias mitigation baselines, but we also maintain and can even improve accuracy. 
The joint improvement of both fairness and prediction accuracy is rare for fairness methods, and we aim to explore this direction further in future work. 
Moreover, we can extend beyond linear mixup to consider other data-driven mixup schemes such as convex clustering~\cite{navarroGraphmadGraphMixup2023}.

\bibliographystyle{ieeetr}
\bibliography{citations}

\begin{thebibliography}{10}

\bibitem{Pessach:2022}
D.~Pessach and E.~Shmueli, ``A review on fairness in machine learning,'' {\em
  ACM Computing Surveys (CSUR)}, vol.~55, no.~3, 2022.

\bibitem{Kamiran:2012}
F.~Kamiran and T.~Calders, ``Data preprocessing techniques for classification
  without discrimination,'' {\em Knowledge and Information Systems}, vol.~33,
  pp.~1--33, 2012.

\bibitem{Feldman:2015}
M.~Feldman, S.~A. Friedler, J.~Moeller, C.~Scheidegger, and
  S.~Venkatasubramanian, ``Certifying and removing disparate impact,'' in {\em
  Intl. Conf. on Knowledge Discovery and Data Mining (SIGKDD)}, pp.~259--268,
  2015.

\bibitem{Calmon:2017}
F.~P. Calmon, D.~Wei, B.~Vinzamuri, K.~N. Ramamurthy, and K.~R. Varshney,
  ``Optimized pre-processing for discrimination prevention,'' in {\em Advances
  in Neural Information Processing Systems}, pp.~3995--4004, 2017.

\bibitem{Zemel:2013}
R.~Zemel, Y.~Wu, K.~Swersky, T.~Pitassi, and C.~Dwork, ``Learning fair
  representations,'' in {\em Intl. Conf. on Machine Learning (ICML)},
  pp.~325--333, 2013.

\bibitem{Donini:2018}
M.~Donini, L.~Oneto, S.~Ben-David, J.~S. Shawe-Taylor, and M.~Pontil,
  ``Empirical risk minimization under fairness constraints,'' in {\em Advances
  in Neural Information Processing Systems}, vol.~31, 2018.

\bibitem{Do:2022}
H.~Do, P.~Putzel, A.~S. Martin, P.~Smyth, and J.~Zhong, ``Fair generalized
  linear models with a convex penalty,'' in {\em Intl. Conf. on Machine
  Learning (ICML)}, vol.~162, pp.~5286--5308, PMLR, 2022.

\bibitem{cotter:2019}
A.~Cotter, M.~Gupta, H.~Jiang, N.~Srebro, K.~Sridharan, S.~Wang, B.~Woodworth,
  and S.~You, ``Training well-generalizing classifiers for fairness metrics and
  other data-dependent constraints,'' in {\em Intl. Conf. on Machine Learning
  (ICML)}, vol.~97, pp.~1397--1405, PMLR, 2019.

\bibitem{poggio:2017}
T.~Poggio, K.~Kawaguchi, Q.~Liao, B.~Miranda, L.~Rosasco, X.~Boix, J.~Hidary,
  and H.~Mhaskar, ``Theory of deep learning {III}: Explaining the
  non-overfitting puzzle,'' {\em arXiv preprint arXiv:1801.00173}, 2017.

\bibitem{zhang_mixup:2018}
H.~Zhang, M.~Cisse, Y.~N. Dauphin, and D.~Lopez-Paz, ``mixup: Beyond empirical
  risk minimization,'' in {\em Intl. Conf. on Learning Representations (ICLR)},
  2018.

\bibitem{zhang2022when}
L.~Zhang, Z.~Deng, K.~Kawaguchi, and J.~Zou, ``When and how mixup improves
  calibration,'' in {\em Intl. Conf. on Machine Learning (ICML)}, vol.~162,
  pp.~26135--26160, PMLR, 2022.

\bibitem{han2022umix}
Z.~Han, Z.~Liang, F.~Yang, L.~Liu, L.~Li, Y.~Bian, P.~Zhao, B.~Wu, C.~Zhang,
  and J.~Yao, ``{UMIX}: Improving importance weighting for subpopulation shift
  via uncertainty-aware mixup,'' in {\em Advances in Neural Information
  Processing Systems}, vol.~35, pp.~37704--37717, 2022.

\bibitem{chuang:2021}
C.-Y. Chuang and Y.~Mroueh, ``Fair mixup: Fairness via interpolation,'' in {\em
  Intl. Conf. on Learning Representations (ICLR)}, 2021.

\bibitem{sharma2020data}
S.~Sharma, Y.~Zhang, J.~M. R\'{\i}os~Aliaga, D.~Bouneffouf, V.~Muthusamy, and
  K.~R. Varshney, ``Data augmentation for discrimination prevention and bias
  disambiguation,'' in {\em AAAI/ACM Conf. on AI, Ethics, and Society},
  p.~358–364, 2020.

\bibitem{Hardt:2016}
M.~Hardt, E.~Price, and N.~Srebron, ``Equality of opportunity in supervised
  learning,'' in {\em Advances in Neural Information Processing Systems},
  vol.~29, 2016.

\bibitem{yao2022cmixup}
H.~Yao, Y.~Wang, L.~Zhang, J.~Y. Zou, and C.~Finn, ``{C-Mixup}: Improving
  generalization in regression,'' {\em Advances in Neural Information
  Processing Systems}, vol.~35, pp.~3361--3376, 2022.

\bibitem{verma2019manifold}
V.~Verma, A.~Lamb, C.~Beckham, A.~Najafi, I.~Mitliagkas, D.~Lopez-Paz, and
  Y.~Bengio, ``Manifold mixup: Better representations by interpolating hidden
  states,'' in {\em Intl. Conf. on Machine Learning (ICML)}, vol.~97,
  pp.~6438--6447, PMLR, 2019.

\bibitem{yun2019cutmix}
S.~Yun, D.~Han, S.~J. Oh, S.~Chun, J.~Choe, and Y.~Yoo, ``{CutMix}:
  Regularization strategy to train strong classifiers with localizable
  features,'' in {\em IEEE/CVF Intl. Conf. on Computer Vision (ICCV)}, October
  2019.

\bibitem{little2022to}
C.~O. Little, M.~Weylandt, and G.~I. Allen, ``To the fairness frontier and
  beyond: Identifying, quantifying, and optimizing the fairness-accuracy
  {Pareto} frontier,'' {\em arXiv preprint arXiv:2206.00074}, 2022.

\bibitem{law_data}
L.~Whiteman, ``The scale and effects of admissions preferences in higher
  education ({SEAPHE}),'' 1998.

\bibitem{navarroGraphmadGraphMixup2023}
M.~Navarro and S.~Segarra, ``{GraphMAD}: Graph mixup for data augmentation
  using data-driven convex clustering,'' in {\em IEEE Intl. Conf. Acoust.,
  Speech and Signal Process. (ICASSP)}, pp.~1--5, 2023.

\end{thebibliography}

\end{document}